\documentclass{article}

\usepackage{arxiv}

\usepackage[utf8]{inputenc} 
\usepackage[T1]{fontenc}    
\usepackage{hyperref}       
\usepackage{url}            
\usepackage{booktabs}       
\usepackage{amsfonts}       		
\usepackage{graphicx}
\usepackage{amsmath}

\title{A Continuous and Interpretable Morphometric for Robust Quantification of Dynamic Biological Shapes}
\date{}

\author{
Roua Rouatbi$^{1,2,3,4}$,
Juan-Esteban Suarez Cardona$^{6,7}$,
Alba Villaronga-Luque$^{2}$,
Jesse V. Veenvliet$^{2,3,5}$, \And
Ivo F.~Sbalzarini$^{1,2,3,4,5}$\\
\\
$^1$ Dresden University of Technology, Faculty of Computer Science \\
$^2$ Max Planck Institute of Molecular Cell Biology and Genetics \\
$^3$ Center for Systems Biology Dresden \\
$^4$ Center for Scalable Data Analytics and Artificial Intelligence (ScaDS.AI) Dresden/Leipzig \\
$^5$ Cluster of Excellence Physics of Life, Technische Universität Dresden, Germany \\
$^6$ Ludwig-Maximilians-Universit\"at M\"unchen \\
$^7$ Munich Center for Machine Learning (MCML)
}

\renewcommand{\shorttitle}{A Continuous and Robust Morphometric}

\hypersetup{
pdftitle={PF-SDMorphometric},
pdfauthor={Roua Rouatbi, Juan-Esteban Suarez Cardona, Alba Villaronga-Luque, Jesse V. Veenvliet, Ivo F.~Sbalzarini},
pdfkeywords={geometric shape analysis, level-set methods, shape quantification, morphometric, vector embedding},
}

\begin{document}

\renewcommand{\thefootnote}{}
\footnotetext{This work has been submitted to the IEEE for possible publication. Copyright may be transferred without notice, after which this version may no longer be accessible.}
\renewcommand{\thefootnote}{\arabic{footnote}}
\maketitle

\begin{abstract}
We introduce the Push-Forward Signed Distance Morphometric (PF-SDM) for shape quantification in biomedical imaging. The PF-SDM compactly encodes geometric and topological properties of closed shapes, including their skeleton and symmetries. This provides robust and interpretable features for shape comparison and machine learning. The PF-SDM is mathematically smooth, providing access to gradients and differential-geometric quantities. It also extends to temporal dynamics and allows fusing spatial intensity distributions, such as genetic markers, with shape dynamics. We present the PF-SDM theory, benchmark it on synthetic data, and apply it to predicting body-axis formation in mouse gastruloids, outperforming a CNN baseline in both accuracy and speed.
\end{abstract}


\section{Introduction}
\label{sec:intro}
The human visual system excels at recognizing and grouping similar-looking objects based on their shape, forming equivalence classes that remain invariant under shape-preserving transformations---rotation, translation, reflection, and scaling. 
%For example, the outline of a flower may be perceived as more similar to a star than to a circle. 
Capturing this in a quantitative metric of shape similarity, however, remains a  challenge in biomedical imaging.

Shape metrics, often called \emph{morphometrics} in biomedical imaging, have been extensively investigated to quantify differences in shapes. Classic morphometrics, such as Generalized Procrustes Analysis (GPA) \cite{Gower1975} and Elliptical Fourier Analysis (EFA) \cite{Kuhl1982}, are widely used but have important limitations: GPA depends on manually selected landmarks, while EFA is sensitive to the starting point and yields correlated features. %\cite{Klein2017}. 
Topological approaches like the Euler Characteristic Transform (ECT) and Persistent Homology Transform (PHT) \cite{Turner2014} capture global mutli-scale features but lack smooth geometric information.

Approaches based on Signed Distance Functions (SDF) \cite{Osher2003-ga} have been introduced to provide geometric representations from which interpretable shape descriptors can be derived. The SDF measures the signed orthogonal distance from any point in the domain to the boundary of a closed shape.
The singularities in the SDF characterize the topological skeleton of the shape, which is important for shape similarity \cite{Torsello2004}. However, most SDF methods are limited to discrete representations of the SDF, and the SDF is not invariant to shape-preserving transformations.

Here, we provide a shape-preserving and continuous SDF variant, the Push-Forward Signed Distance Function (PF-SDF). It is mathematically smooth and relies on a common reference domain for shape invariance. From the Fourier transform of the PF-SDF, we derive the Push-Forward Signed Distance Morphometric (PF-SDM) to capture topological and geometric shape features. We show that this morphometric yields robust and interpretable embeddings that extend to temporal dynamics and can be fused with spatial distributions of, e.g., genetic markers. We showcase this by applying PF-SDM to predicting body-axis formation in mouse gastruloids. Formulating the resulting machine-learning problem in PF-SDF space, instead of pixel space, imrpoves both accuracy and computational cost over a CNN baseline.

\section{Method}
The PF-SDM is computed in three stages: SDF computation, PF-SDF mapping, and morphometric evaluation. 

We define a shape $S\subseteq \Omega$ as a closed smooth manifold of co-dimension one~\cite{M2010-sj}, embedded in a domain $\Omega:=(-1,1)^2$, and $\mathbb{S}_M:=\{S_i\}_{i=1}^N\subseteq \Omega$ a set of $N\in \mathbb{N}$ shapes. We consider a data set $I_N \subseteq \mathbb{R}^{p\times p}$ of 2D $p\times p$ images. Each image is preprocessed by segmenting the shape and scaling it to fit within the unit circle, yielding a set of points $\mathbf{x}_{S_i}:=\{x_{l,i}\}_{l=1}^{N}\subseteq S_i$ sampled on each shape $S_i$.

\subsection{Signed Distance Function computation}\label{sub:pip_1}
We first compute the classic Signed Distance Function (SDF) $\phi_S: \Omega \mapsto \mathbb{R}$ for a given shape $S$ by numerically solving the viscous Eikonal equation~\cite{evans10}. The adjustable viscosity term $\nu\in \mathbb{R}_+$ ensures numerical stability by regularizing the SDF around singularities. Specifically, we solve
\begin{equation}\label{eq:eik}
    \left\{\begin{aligned}
        \|\nabla\phi(x,y)\|_2^2+\nu \Delta\phi(x,y) = 1 &\qquad (x,y)\in \Omega ,\\
        \phi(x,y)  = 0 &\qquad(x,y)\in S,
    \end{aligned}\right.
\end{equation}
where $\|\cdot\|_2$ denotes the $\ell^2$-norm of a vector. The boundary condition for Eq.~\eqref{eq:eik} is sampled at the points $\mathbf{x}_S\subseteq S$ from the preprocessing step. 
Due to the $\ell^2$-term, Eq.~\eqref{eq:eik} is nonlinear in $\phi$. We solve it numerically using polynomial surrogate models (PSM) and Sobolev cubatures \cite{Neg_Sob_cardona}. This results in a smooth polynomial approximation of the SDF.
%\cite{Katzourakis2014-fe}

\subsection{Push-Forward Signed Distance Function}\label{sub:pip_2}

The SDF $\phi_S$ provides a complete representation of a shape $S$~\cite{Delfour1994}. While SDFs are widely used in shape analysis, they are not invariant to shape-preserving transformations. To address this limitation, we introduce the {\em push-forward signed distance function} (PF-SDF) $\phi_{S^*} : \Omega_r \mapsto \mathbb{R}$. This achieves scale and translation invariance by mapping to a reference domain $\Omega_r \subseteq \mathbb{R}^2$, deforming the original SDF $\phi_S$ such that its zero level set aligns with the reference shape $S_r = \partial\Omega_r$, i.e., $S_r = \{x \in \mathbb{R}^2 : \phi_{S^*}(x) = 0\}$.

In the following, we consider $\Omega_r\subseteq \mathbb{R}^2$ to be the unit disk $\Omega_r := \{x\in \mathbb{R}^2: \|x\|_2\leq 1\}$ with boundary $S_r = \{x\in \mathbb{R}^2: \|x\|_2= 1\}$. The {\em push-forward map} of the SDF $\phi$ onto $S_r$ is represented as a deformation map $\Psi_{\zeta}:S \rightarrow \mathbb{R}^2$, deforming $S$ to $S_r$ as $\Psi_\zeta(x,y):=(x+\nu_{\zeta_x}(x,y), \, y+\nu_{\zeta_y}(x,y))$. Here, $\nu_{\zeta_x}, \nu_{\zeta_y}\in \Pi_{m}(S)$ are polynomials of degree $m\in \mathbb{N}$, parametrized by their coefficients $\zeta_x,\zeta_y\in \mathbb{R}^{n+1}$ in the Chebyshev basis by solving
\begin{align}\label{eq:def_map}
&\min \limits_{\zeta_x,\zeta_y\in \mathbb{R}^{n+1}}\int\limits_{S}(\Psi_\zeta - \mathbf{x}_r)^2\, \mathrm{d}s\qquad\textrm{ with }\\&
\mathbf{x}_r(x,y):=\arg\min\limits_{x_r,y_r\in S_r}\|x_r-x\|_2+\|y_r-y\|_2 \, .\notag
\end{align}
This determines $\Psi$ on the subset $S \subseteq \Omega$, as the objective functional is integrated over $S$ rather than the entire domain $\Omega$. To obtain a global deformation map $\overline{\Psi}_{\zeta} : \Omega \to \mathbb{R}^2$, we construct a harmonic extension of $\Psi$~\cite{Shi2015-xq}. In polar coordinates $(\theta, r)$, this yields:
\begin{equation}\label{eq:PF_SDF}
    \phi_{S^*}(\theta, r)=\phi_S(\overline{\Psi}_{\zeta_x}(x(\theta,r), y(\theta,r)),\overline{\Psi}_{\zeta_y}(x(\theta,r),y(\theta,r))
\end{equation}
with $x(\theta,r)=r\cos(\theta)$, $y(\theta,r)=r\sin(\theta)$.

\subsection{Morphometric evaluation}\label{sub:pip_3}
The PF-SDF maps any shape to a common reference, where $r=1$ is the shape boundary and $r=0$ its center, while retaining the geometric richness of traditional SDF formulations. In particular, morphometrics can be defined as shape similarity measures between the PF-SDFs $\phi_{S_i^*},\phi_{S_j^*}$ of two shapes $S_i, S_j$. Here, we propose a morphometric based on the Fourier transform of the PF-SDF. 

The Fourier transform is a natural choice because the PF-SDF for a given radius $r$, $\theta\mapsto \phi_{S^*}(\theta,\cdot)$, is $2\pi$-periodic. Therefore, we consider 
\begin{equation}
    \hat{\phi}_{S^*}(\theta,r)=\sum\limits_{n=0}^{N_F} c_n(r)\, \mathrm{e}^{-i n\theta},
\end{equation}
where $N_F\in\mathbb{N}$ is a fixed number of Fourier coefficients $c_n:(0,1)\rightarrow \mathbb{R}$. The coefficients are functions of $r$: 
\begin{equation}
    c_n(r)=\frac{1}{2\pi}\int\limits_{0}^{2\pi}\phi_{S^*}(\theta,r)\, \mathrm{e}^{-in \theta}\,\mathrm{d}\theta \, . 
\end{equation}

This renders the PF-SDM invariant to rotation and reflection. The map $S\mapsto \mathfrak{c}_{\phi_S}(r):=\{c_n(r)/\|c_n(r)\|_2\}_{n=1}^{N_F}$, $\mathfrak{c}_{\phi_{S}}:(0,1)\rightarrow\mathbb{R}^{N_F}$, directly provides a vector embedding of the shape $S$.

The PF-SDF can optionally account for an intensity field $I_S:\Omega\rightarrow\mathbb{R}$. This can be any scalar field in the image, for example the fluorescence intensity of a genetic marker. Using the deformation map from Eq.~\eqref{eq:def_map}, the intensity signal is mapped to the same unit disk, yielding $I_{S^*}:(0,2\pi)\times(0,1)\rightarrow\mathbb{R}$ with normalized Fourier coefficients  $\mathfrak{c}_{I_{S}}$.

Together, this defines the intensity-aware PF-SDM distance between two shapes $S_1,S_2\subseteq\Omega$,
\begin{align}
    d_{\phi}(S_1, S_2):=& \int\limits_{0}^1\bigr({\mathfrak{c}}_{\phi_{S_1}}(r)-{\mathfrak{c}}_{\phi_{S_2}}(r))^2\, \mathrm{d}r  \notag \\& +\int\limits_{0}^1\bigr({\mathfrak{c}}_{I_{S_1}}(r)-{\mathfrak{c}}_{I_{S_2}}(r))^2\, \mathrm{d}r\, . \label{eq:distance formula}
\end{align}
The PF-SDM of any shape is invariant to shape-preserving transformations, and so is the above distance.

\section{Results}
\label{sec:results}

We validate the PF-SDM on a synthetic benchmark data set to confirm its invariance properties and compare it with previous morphometrics in terms of its robustness and interpretability. Then, we illustrate the application of the PF-SDM to a biological problem that includes temporal dynamics of shapes and of the intensity field of a genetic marker. Specifically, we consider the problem of predicting body-axis formation in mouse gastruloids from fluorescence microscopy images. 

\subsection{Synthetic Validation Benchmark}

\begin{figure}[htb]
\begin{minipage}[b]{1.0\linewidth}
  \centering
  \centerline{\includegraphics[width=14cm]{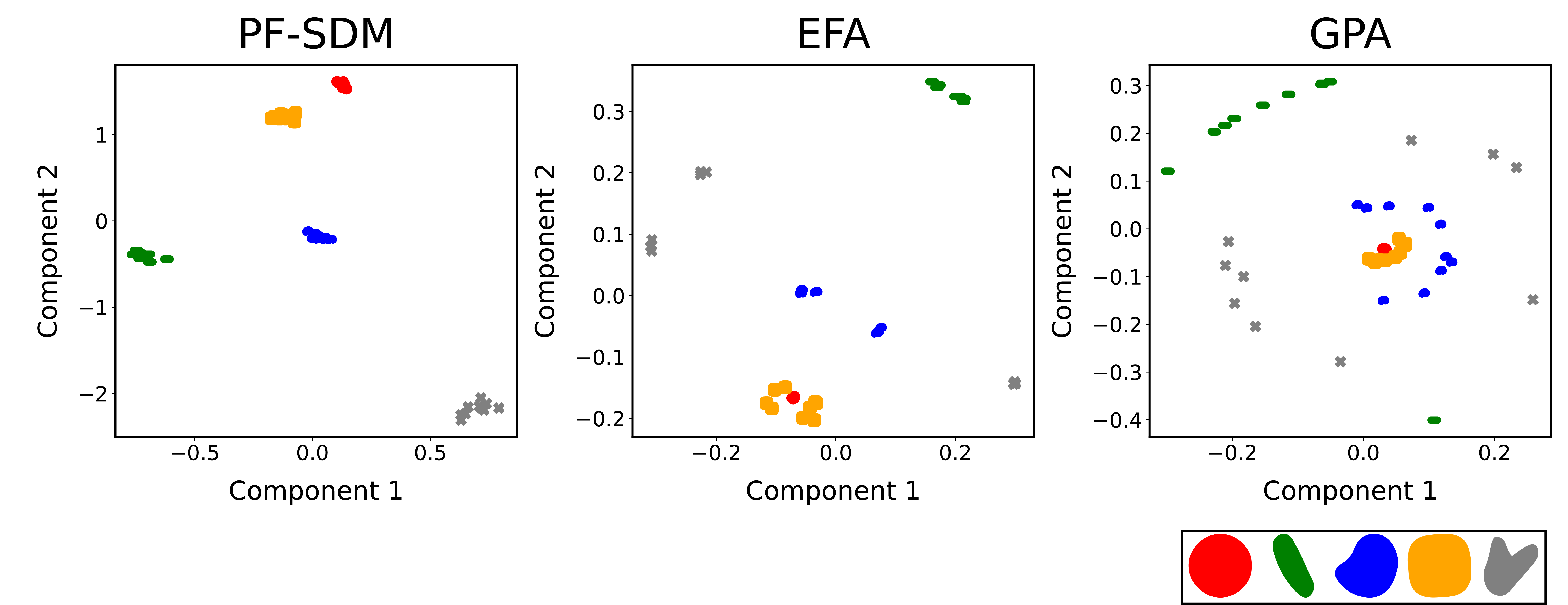}}
  \caption{First two MDS components of the shape distance matrices computed with PF-SDM (left), EFA (center), and GPA (right) over ten shape-preserving transformations of each of the five test shapes shown in the inset legend. Point colors match shape-legend colors.}
  \label{fig:clustering}
\end{minipage} 
\end{figure}

We assess the robustness of the PF-SDM to shape-preserving transformations by considering a synthetic data set of five base shapes: a circle $(S_1)$, an elongated shape $(S_2)$, an asymmetrically folded shape $(S_3)$, a rounded square $(S_4)$, and a branched shape $(S_5)$ (for examples see Fig.~\ref{fig:clustering}, bottom-right). Each base shape is transformed ten times by random translation, rotation, reflection, and isotropic scaling, resulting in 50 transformed shapes. We compute complete pairwise distance matrices between these 50 shapes using three different morphometrics:
\begin{itemize}
\item \textbf{PF-SDM}: Shape distances are computed using Eq.~(\ref{eq:distance formula}) with $N_F = 10$ Fourier coefficients. 
%%\item \textbf{EFA}: We compute the coefficients $\mathfrak{c}:=\{\mathfrak{c}_{S_1},\ldots , \mathfrak{c}_{S_{50}}\}$ of the EFA using the \texttt{spatial-efd} package \cite{WDGrieve2017}. The coefficients are normalized to ensure translation, rotation, and scale invariance. Distances are computed using the first 10 harmonics, $d_\textrm{EFA}:=(\|\mathfrak{c}_{S_i}-\mathfrak{c}_{S_j}\|_2)_{i,j=1}^{50}$.
\item \textbf{EFA}: We represent each shape using the first 10 elliptical Fourier harmonics, computed with the \texttt{spatial-efd} package \cite{WDGrieve2017}. The resulting coefficients are normalized, yielding one coefficient vector $\mathfrak{c}_{S_i}$ per shape.
%Pairwise Euclidean distances between the 50 vectors are then computed to form the distance matrix  $d_\textrm{EFA}:=(\|\mathfrak{c}_{S_i}-\mathfrak{c}_{S_j}\|_2)_{i,j=1}^{50}$.
\item \textbf{GPA}: We sample equidistant boundary points from EFA. GPA is then used on these landmark points to compute pairwise distances between aligned shapes.
\end{itemize}
The first two components of a Multidimensional Scaling (MDS)~\cite{Cox2000} reduction of the resulting distance matrices are shown in Fig.~\ref{fig:clustering}.

The data shows that PF-SDM yields the most robust clustering under shape-preserving  transformations. Randomly perturbed instances of the same shape remain more tightly grouped than in  EFA and GPA. In addition, the PF-SDM uncovers a meaningful hierarchy of shape similarity. The circle and rounded square clusters are closest, followed by the folded shape, while the elongated and branched shapes are mapped farthest apart. This demonstrates the robustness to transformations and geometric interpretability of PF-SDM and its superior performance over EFA and GPA.

Interestingly, the hierarchy revealed by PF-SDM---from spherical to folded to branched---mirrors the typical morphological progressions observed in developmental biological systems.

\subsection{Application to Mouse Gastruloids}

To illustrate the use of the PF-SDM in a challenging developmental biology application, we consider the developmental dynamics of mouse gastruloids. Gastruloids are an \textit{in vitro} stem-cell-based embryo model recapitulating aspects of body patterning and morphogenesis \cite{ beccari_2018}. Their generation induces expression of the transcription factor Brachyury, followed by symmetry breaking and body-axis formation \cite{beccari_2018, villarongaluque_2025}. During this morphogenetic process, gastruloids transition from a spherical to an ovoid/teardrop and finally to an elongated shape \cite{beccari_2018, villarongaluque_2025}. Unlike a natural embryo, however, gastruloids occasionally form multiple body axes \cite{bennabi_2024}. We analyze time-lapse videos of gastruloids with an mCherry-labeled Brachyury reporter \cite{veenvliet_2020, villarongaluque_2025}, imaged in brightfield and mCherry channels every 45\,min. The data set comprises videos of 78 gastruloids, 65 developed a single axis and 13  multiple axes.

\begin{figure}[htb]
\begin{minipage}[b]{1.0\linewidth}
  \centering
  \centerline{\includegraphics[width=10 cm]{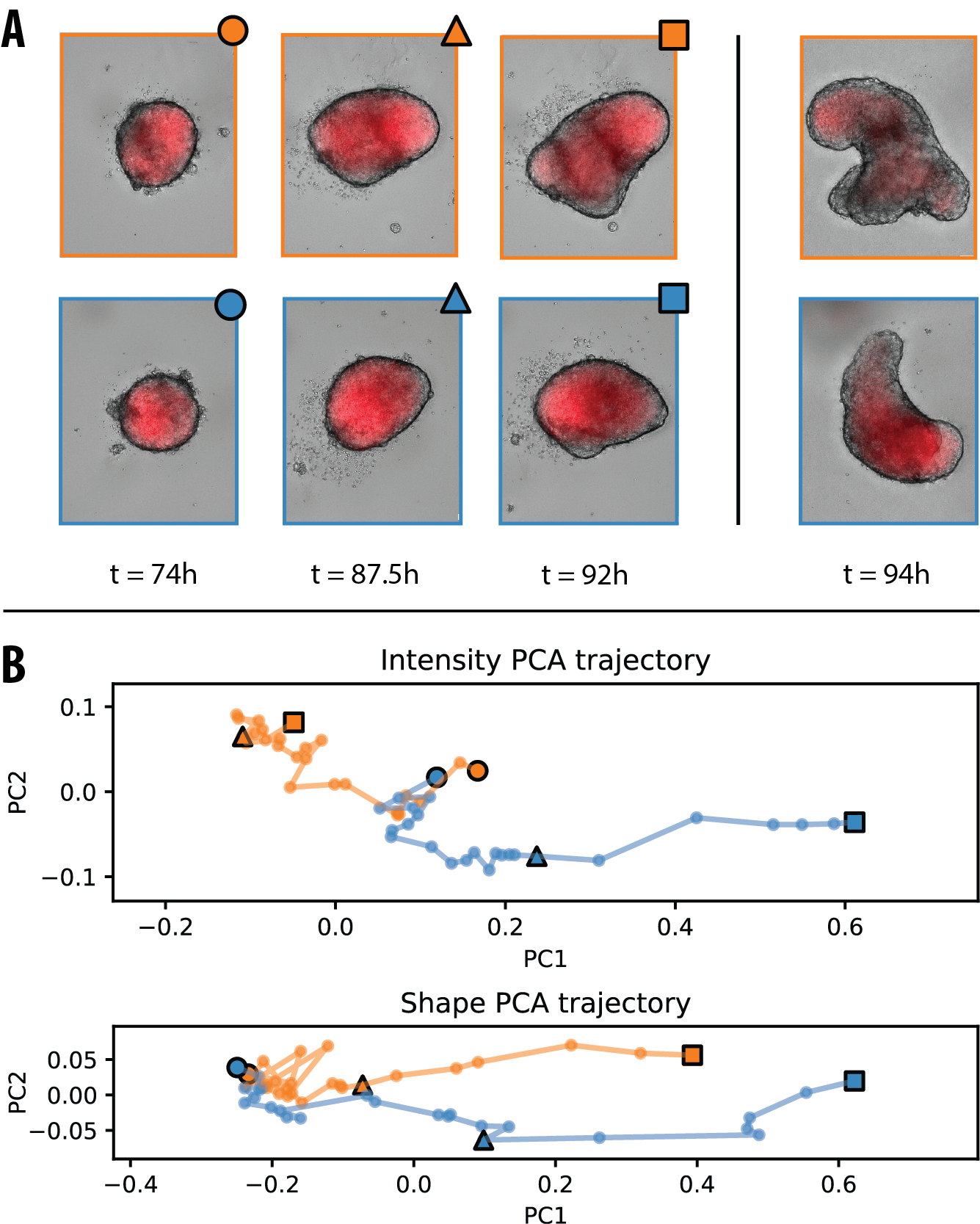}}
  
  \caption{Shape analysis of developing mouse gastruloids. {\bf (A)} Representative gastruloid images from the single-axis (blue) and multi-axes (orange) classes at early (circle), intermediate (triangle), and late (square) times, along with the  posterior reference at 94\,h. Brachyury::mCherry intensity is shown in red.
  {\bf (B)} PCA trajectories of $N_F=25$ PF-SDF Fourier coefficients $\mathfrak{c}$ for Brachyury intensity (top) and shape (bottom) for the examples from (A) across 25 time points.}
  \label{fig:gastruloids data}
\end{minipage} 
\end{figure}

We represent each gastruloid independently at every time point using the PF-SDF
for its shape and the Brachyury intensity field. 
Figure~\ref{fig:gastruloids data}A shows representative instances from the two gastruloid classes across time. Figure~\ref{fig:gastruloids data}B shows the PCA trajectories of $N_F=25$ Fourier coefficients of the PF-SDF for the examples from (A). At early stages (circle), gastruloids exhibit similar morphologies and intensity distributions. At intermediate stages (triangle), shape differences between the two classes start to be captured by the PF-SDF (Fig.~\ref{fig:gastruloids data}B). The differences in shape persist until the final time, whereas differences in shape-corrected intensity further amplify, indicating that molecular asymmetry precedes shape dynamics.

We next ask whether the eventual formation of multiple axes can be predicted from images at earlier times (74$\ldots$92\,h) when morphological differences between the classes are not yet obvious to the human eye (Fig.~\ref{fig:gastruloids data}A). This makes the task intrinsically challenging but biologically meaningful, since it tests whether early cues in shape and gene expression can predict future developmental fate. Ground-truth labels (single-axis vs.~multi-axes) for training were assigned retrospectively, based on morphological outcomes observed at later times (94$\ldots$120\,h), when the emergence of one or multiple axes becomes unambiguous.

For classification, the PF-SDF Fourier coefficients for each time point are concatenated into one vector. We train logistic regression classifiers with balanced class weights separately for the shape and intensity features and compare classification performance with a convolutional neural network (CNN) baseline~\cite{villarongaluque_2025} trained directly on pixel segmentation images and raw intensity images, using data augmentation tailored to each (affine transformations for shapes, affine and noise perturbations for intensities).
All classifiers output separate class probabilities for the shape and intensity channels, $p_{\text{SDF}}$ and $p_{\text{I}}$. Final predictions are computed by late fusion as $p_{\text{fusion}} = \alpha \, p_{\text{SDF}} + (1-\alpha) \, p_{\text{I}}$, with $\alpha \in [0,1]$ tuned for each classifier to maximize balanced accuracy on held-out test splits. Evaluation is done with repeated stratified shuffle splits ($k=5$, test size 20\%) \cite{scikit-learn}.

\begin{figure}[htb]
\begin{minipage}[b]{1.0\linewidth}
  \centering
  \centerline{\includegraphics[width=12 cm]{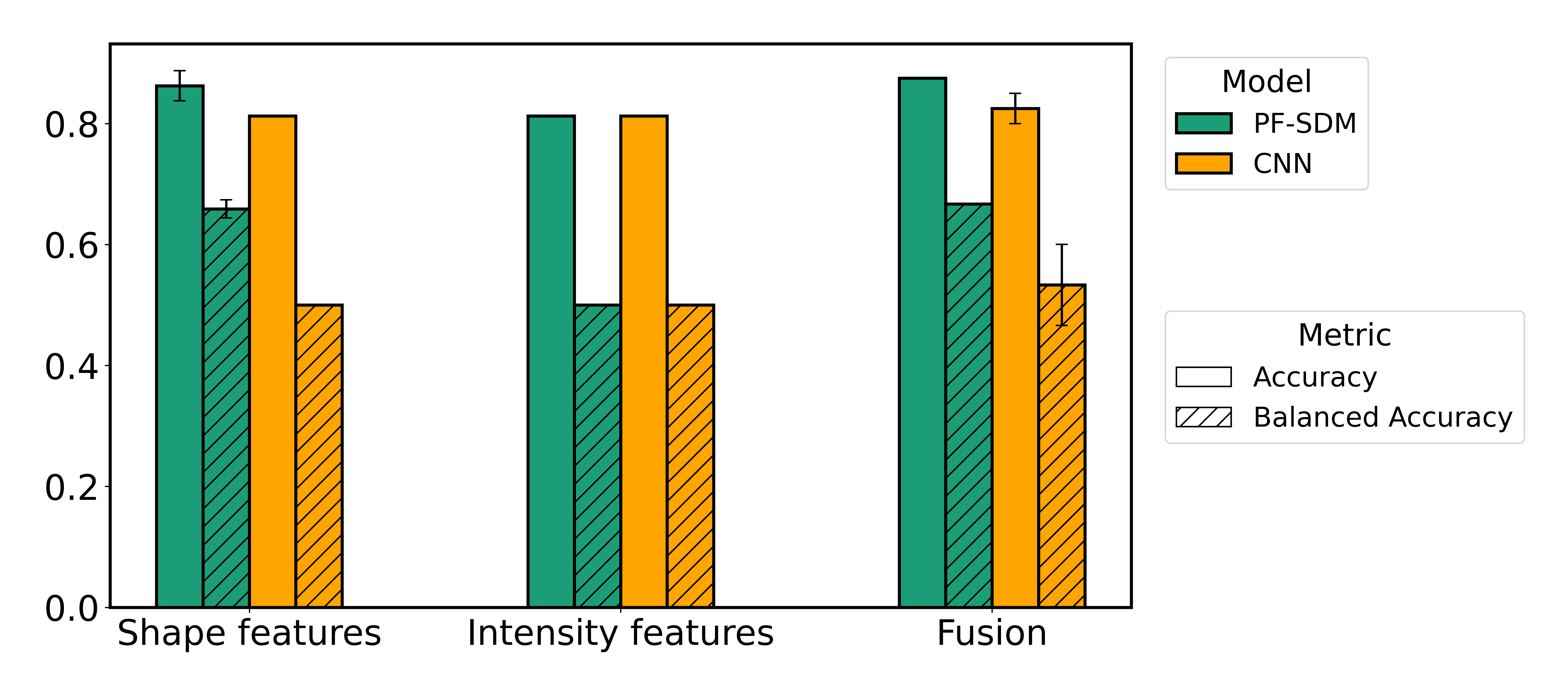}}
  \caption{Comparison of classification performance between logistic regression over PF-SDM features and the CNN baseline~\cite{villarongaluque_2025}. Accuracy (plain) and Balanced Accuracy (shaded) are shown for shape, intensity, and fusion (shape+intensity) features. Error bars: empirical standard deviation across five stratified test splits.}
  \label{fig:gastruloids classification}
\end{minipage} 
\end{figure}

The results in Fig.~\ref{fig:gastruloids classification} show that logistic regression over PF-SDM features consistently outperforms the CNN baseline both in accuracy and balanced accuracy, despite the severe class imbalance. Moreover, classification based on PF-SDM required an order of magnitude less compute resources (4\,min vs.~50\,min for CNN training). The accuracy gains in PF-SDM are explained by the shape features, whereas the intensity features achieve the same performance as the CNN. Importantly, the fusion of shape and intensity improves the mean performance across test splits, suggesting complementary biological contributions of morphology and gene expression.

\section{Conclusion}\label{sec:discussion}

We presented the Push-Forward Signed Distance Morphometric (PF-SDM), a continuous geometric morphometric that is invariant to shape-preserving transformation. We provided the theoretical foundation for PF-SDM and a practical algorithm for its computation (available in Python from \url{https://git.mpi-cbg.de/mosaic/software/machine-learning/pf-sdm}). Through experiments on synthetic shapes, we confirmed that PF-SDM generates robust (to shape transformations) and geometrically interpretable hierarchical embeddings of geometric objects based on their shapes, outperforming both Elliptical Fourier Analysis (EFA) and Generalized Procrustes Analysis (GPA). We then applied the proposed PF-SDM to the biological problem of predicting body-axis formation in mouse gastruloids from microscopy images. There, simple logistic regression over FP-SDM features outperformed a custom CNN trained on the raw images while requiring ten-fold less computer time. This showcases the benefits of formulating a machine-learning question in a more appropriate space with the correct in- and equi-variances. Indeed, the PF-SDM transforms spatiotemporal shape dynamics to mathematically smooth and geometrically interpretable compact features that can directly be fused with gene-expression patterns. 

For the sake of clarity and simplicity, we considered only two-dimensional shapes and their temporal dynamics. Extending the PF-SDF framework to 3D+time is subject to ongoing work. All mathematical concepts do, however, generalize to higher dimensions. More difficult is the generalization from closed shapes to open or non-singly-connected manifolds, which would require tree or vector SDFs~\cite{Xiao2016}. PF-SDM could also be generalized to Sobolev norms \cite{Raa2003-ok}, which could help capture higher-order geometric features.
Finally, future work could attempt to formally prove the differentiability of PF-SDM distances, which could be conjectured from their smoothness and continuity. This would open new avenues for incorporating PF-SDM terms in loss functions of machine-learning models to enhance shape representation \cite{Annafoix}.  

\section{Acknowledgments}
\label{sec:acknowledgments}
This work was supported by the Max Planck Society and the German Federal Ministry of Research, Technology and Space (BMFTR) as part of the Center for Scalable Data Analytics and Artificial Intelligence (ScaDS.AI). JVV acknowledges support from a European Innovation Council (EIC) Pathfinder grant under the Horizon Europe Research and Innovation Program (Horizon-EIC-2021-PathfinderChallenges-01 101071203, SUMO).

\bibliographystyle{unsrtnat}

\end{document}